\documentclass{article}

\usepackage[final]{neurips_2026}
\makeatletter
\providecommand{\@trackname}{}
\makeatother

\usepackage{amsmath, amssymb}
\usepackage{graphicx}
\usepackage{booktabs}
\usepackage{hyperref}
\usepackage{caption}
\usepackage{longtable}
\usepackage{geometry}
\usepackage{enumitem}
\usepackage{makecell}
\usepackage{listings} 
\newcommand{\keywords}[1]{%
  \noindent\textbf{Keywords:} #1
}

\graphicspath{{figures/}}

\title{The Mask Is Not the Model: Auditing Prefix Invariance in Attention, State-Space, and Hybrid Sequence Models}
\author{
Taebong Kim \quad
Youngsik Hong \quad
Minsik Kim \quad
Sunyoung Choi \\
Jaewon Jang \quad
Minseo Kim \\
VIDRAFT AI Research · QuantumOS, Seoul, Republic of Korea \\
\texttt{arxivgpt@gmail.com}
}

\begin{document}
\maketitle

\begin{abstract}

Hybrid sequence models, which combine attention with state-space, recurrent, convolutional, and other sequence mixers, 
have become a standard architecture for efficient long-context language modeling. Yet they rely on a property that is rarely verified in practice: in an autoregressive model, 
representations at position $t$ must not depend on inputs at positions greater than $t$.

We formalize this requirement as \textbf{prefix invariance}: two sequences that differ only at the final token must produce identical representations at every earlier position and every layer. 
We introduce a simple audit requiring only two forward passes and no training, gradients, labels, or accelerators. The audit records a per-layer difference score $\Delta_\ell$ 
and reports the first layer at which it becomes nonzero, turning leak detection into fault localization.

We show that the field's de facto audit procedure---inspection of attention masks---is incomplete by construction. 
Causality is a property of the entire computational graph rather than of any single operator, and future information can leak through scans, aggregations, 
or normalization paths while all attention masks remain correct. Across 192 injected-fault trials on eight public checkpoints, mask inspection detected none.

Across attention-only, state-space, recurrent, and hybrid architectures, our method localized all 192/192 injected faults to the exact layer. 
All audited checkpoints were clean at the tested length, with $\max |\Delta| = 0$, and four models produced bit-identical outputs.

A static audit of chunked-scan implementations in \texttt{transformers}, followed by dynamic testing, uncovered the same causal defect in two released models: 
\texttt{Zamba2} and \texttt{Nemotron-H}. Both leak future information across chunk boundaries due to an axis error in the inter-chunk recurrence 
and become clean after replacement with the reference implementation. The defect is structural and reproducible across seeds.

The finding revealed a second methodological requirement: audit sequences must exceed a model's chunk or window parameter. 
A logits-only perturbation test detects leaks but cannot localize them, while a gradient-based method matches localization at higher computational cost. 
Apparently clean verdicts may also arise from non-responsive model loads, motivating a positive-control gate.

The method fits on a single page, executes in seconds, and requires only forward evaluation. We release audit logs and fault specifications, 
and argue that causal-correctness certificates should accompany future architecture releases.

\end{abstract}


\section{Introduction}

\subsection{A property nobody checks}

Autoregressive language models have become the dominant paradigm for modern sequence modeling
and underpin systems ranging from GPT-style language models to instruction-tuned assistants
\citep{radford2018gpt,radford2019gpt2,brown2020gpt3}.

More broadly, modern foundation models emerged from a long line of sequence-learning research,
including sequence-to-sequence learning
\citep{sutskever2014seq2seq},
attention-based neural machine translation
\citep{bahdanau2015attention},
bidirectional transformer pretraining
\citep{devlin2019bert},
and generalized autoregressive objectives
\citep{yang2019xlnet}.

An autoregressive model is meaningful only if it satisfies one structural constraint:
the output at position $t$ may depend on inputs at positions $\le t$, and on nothing else.

Every practical use of such a model---teacher-forced training, perplexity evaluation, incremental decoding, KV-cache reuse, and speculative decoding---silently assumes this property. 
Yet contemporary model releases routinely report parameter counts, context lengths, training tokens, and benchmark performance while providing no evidence that the released implementation is causally correct.

Historically this omission was less concerning because architectures were relatively homogeneous.
Decoder-only Transformers contain a single dominant temporal mixing mechanism, self-attention,
whose causal behavior is governed by an explicit mask.

The modern self-attention paradigm originates with neural attention mechanisms
\citep{bahdanau2015attention}
and was standardized by the Transformer architecture
\citep{vaswani2017attention}.

Consequently, practitioners often treated inspection of the mask as a de facto audit of causality.

This assumption no longer holds. Recent sequence-model architectures increasingly combine heterogeneous mixers within a single stack, including local attention
\citep{beltagy2020longformer},
sparse attention
\citep{child2019sparse,zaheer2020bigbird},
memory-augmented attention
\citep{gupta2022memorizing},
linear recurrences
\citep{peng2023rwkv,orvieto2023resurrecting},
structured state-space models
\citep{gu2021s4,gu2024mamba},
state-space dual architectures
\citep{dao2024transformers},
recurrent-attention hybrids
\citep{de2024griffin,botev2024recurrentgemma},
and convolutional alternatives
\citep{poli2023hyena}.

More broadly, efficient-sequence-modeling research now spans sparse, local, recurrent,
convolutional, and state-space approaches designed to overcome the quadratic cost of standard
self-attention
\citep{tay2022efficient}.

In such systems, explicit attention masks govern only a subset of layers and may be absent altogether.
The traditional audit procedure therefore stopped covering most of the computation graph.

\subsection{Why silent leakage is the dangerous failure mode}

Causal leakage is not a crash, an exception, or an obvious malfunction. Instead, it is a failure mode that can make development metrics appear \emph{better}.

If future information reaches earlier positions, the next-token prediction task becomes artificially easier. Training loss decreases, 
validation perplexity decreases, and benchmark scores computed from teacher-forced likelihoods may improve, despite the implementation violating the assumptions of autoregressive inference. 
Similar forms of leakage have repeatedly produced misleading conclusions in machine learning research 
and have been identified as a major contributor to reproducibility failures across scientific domains \citep{kapoor2023leakage}.

The practical danger is asymmetry: leakage improves the very metrics used to select models while only becoming apparent during free-running generation, 
deployment, or downstream evaluation. By the time the issue is discovered, architecture comparisons and performance claims may already have been contaminated. 
This observation motivates a structural correctness check that is simple enough to run routinely and inexpensive enough to become part of standard evaluation practice.

\subsection{Motivation: three faults in a row}

This work began as an engineering tool rather than a research project. During development of an internal hybrid sequence-model stack, we encountered three distinct causal failures:

\begin{enumerate}
\item \textbf{A doubled output shift.} Two independently correct alignment operations combined to expose one future token.

\item \textbf{An incorrect causal condition in a custom attention implementation.} A hand-written attention path replaced a causal neighborhood with a symmetric one.

\item \textbf{A differential-attention aggregation fault.} A subtraction between attention maps was performed after a sequence-wide aggregation step had already mixed information across time.
\end{enumerate}

Each fault required substantial manual debugging despite violating the same underlying property. None produced obvious anomalies in conventional training telemetry. 
The recurring pattern suggested that causal correctness should not require bespoke reasoning about individual architectures. 
Instead, it should be testable through a uniform property of the forward computation itself.

This perspective is closely related to the philosophy of metamorphic testing, where correctness is verified through invariance relations 
rather than task-specific labels \citep{xie2011metamorphic}. 
Similar ideas have successfully exposed defects in deep-learning systems through automated testing frameworks such as 
DeepXplore and DeepTest \citep{pei2017deepxplore,tian2018deeptest}. Our position is that autoregressive language models admit a particularly natural metamorphic relation: 
agreement of all prefix representations under perturbations that occur strictly in the future. 
Unlike conventional testing settings, the oracle is intrinsic to the definition of autoregressive causality itself.

The broader motivation is not confined to language models.
Sequence modeling now spans language, speech, vision, and time-series domains,
including WaveNet-style autoregressive audio generation
\citep{oord2016wavenet},
Vision Transformers
\citep{dosovitskiy2021vit},
and deep time-series models
\citep{fawaz2019deep}.
A general-purpose correctness audit therefore has potential value beyond a single architecture family.

\subsection{Contributions}

Our contributions are:

\begin{enumerate}

\item \textbf{A named property and a test.}
We formalize \emph{prefix invariance} and introduce a two-forward-pass, gradient-free audit that identifies not only whether a causality violation exists but also the first layer where it appears.

\item \textbf{An incompleteness argument.}
We show that mask inspection is not a sound test for causal correctness because causality is a property of the entire computational graph rather than of any single operator.

\item \textbf{A taxonomy of temporal leakage.}
We organize temporal leakage mechanisms into four classes spanning eight executable fault patterns.

\item \textbf{A systematic comparison against existing practices.}
We compare our approach against logits-only perturbation checks, shuffled-suffix evaluation, cache-consistency tests, and gradient-based localization methods.

\item \textbf{A cross-architecture audit.}
We evaluate attention-only, state-space, recurrent, and hybrid architectures,
including models derived from Transformer
\citep{vaswani2017attention},
RWKV
\citep{peng2023rwkv},
LRU-style recurrent networks
\citep{orvieto2023resurrecting},
Mamba
\citep{gu2024mamba},
state-space dual formulations
\citep{dao2024transformers},
and Griffin/RecurrentGemma hybrids
\citep{de2024griffin,botev2024recurrentgemma}.

\item \textbf{A methodological lesson from audit failure.}
We show that apparently clean results are uninterpretable unless accompanied by a positive-control demonstration on the same loaded checkpoint.

\item \textbf{A code-lineage audit of released models.}
A static analysis of chunked scan implementations predicted causal violations in two released hybrid models, and dynamic audits confirmed both predictions.

\item \textbf{A second auditing norm.}
We demonstrate that audit sequence length must exceed architecture-specific chunk, kernel, or window parameters to exercise the relevant execution paths.

\item \textbf{A release recommendation.}
We argue that causal-correctness certificates should accompany architecture releases in the same way that parameter counts, benchmark results, and context lengths are routinely reported.

\end{enumerate}

Our approach is conceptually related to three strands of work:
metamorphic testing
\citep{xie2011metamorphic},
automated testing of deep-learning systems
\citep{pei2017deepxplore,tian2018deeptest},
and reproducibility auditing
\citep{kapoor2023leakage,dacrema2019really,dacrema2021troubling}.

Unlike these approaches, we evaluate a deterministic semantic property whose oracle is derived
directly from the autoregressive model definition.

\section{Method}

\subsection{Definition}

Let $f$ be a model mapping a token sequence $\mathbf{x} \in V^T$ to per-layer representations $h_\ell(\mathbf{x}) \in \mathbb{R}^{T\times d}$ for $\ell = 1,\ldots,L$, 
and to output logits.

\begin{quote} 
  \textbf{Prefix invariance.} For all $\mathbf{x}, \mathbf{x}^{\prime} \in V^T$ and all $t < T$, \[ \mathbf{x}[:t] = \mathbf{x}^{\prime}[:t] \;\Rightarrow\; 
  h_\ell(\mathbf{x})[t] = h_\ell(\mathbf{x}^{\prime})[t] \] for every layer $\ell$. 
\end{quote}

The single-token instance is sufficient to expose any violation and is the cheapest to construct: take \textbf{x¹} and \textbf{x²} identical except at the final position \textit{$T-1$}.

\subsection{The test}

\begin{quote}
\small
\ttfamily

INPUT \quad model $f$ with layer list $[1..L]$; sequence length $T$; threshold $\tau$

OUTPUT \quad verdict $\in \{\mathrm{CLEAN}, \mathrm{LEAK}\}$; first offending layer $\ell^*$

\vspace{0.5em}

1.\ $x^1 \leftarrow$ random token sequence of length $T$

2.\ $x^2 \leftarrow$ copy of $x^1$ with position $T-1$ replaced by a different token

3.\ attach a forward hook to every layer, capturing its output

4.\ $h^1 \leftarrow f(x^1)$ with caching disabled \hfill \# forward pass 1

5.\ $h^2 \leftarrow f(x^2)$ with caching disabled \hfill \# forward pass 2

6.\ for $\ell = 1..L$:

\hspace*{1.5em}
$\Delta_\ell \leftarrow
\max \left|
h^1_\ell[0:T-1]-h^2_\ell[0:T-1]
\right|$
\hfill \# exclude the differing position

7.\ $\ell^* \leftarrow \min \{\ell : \Delta_\ell > \tau\}$, or NONE

8.\ return $(\mathrm{LEAK}, \ell^*)$ if $\ell^*$ exists else $(\mathrm{CLEAN}, -)$

\end{quote}

That is the entire method. It is printed here in full deliberately: the argument of this paper is that the check is cheap enough to be mandatory, and a check that does not fit on a page is not.

\subsection{Four design choices, and why each matters}

\textbf{(1) Per-layer hooks rather than output logits.} Reading only the final logits answers \textit{whether} the model leaks. 
It cannot answer \textit{where}, because every layer's contribution has been mixed by the output projection. 
Since the practical purpose of the test is to send an engineer to a specific line of code, the layer index is the deliverable. 
§3.3 quantifies this: the logits-only variant detects every injected fault and localizes none.

\textbf{(2) Excluding the final position from the comparison.} Position $T-1$ legitimately differs---that is the perturbation. 
Comparing positions $[0, T-1)$ isolates exactly the prefix that must be invariant.

\textbf{(3) Disabling caching during the test.} With incremental caching enabled, the two passes may take different code paths and reuse state across
 calls, 
so a difference (or its absence) would no longer be attributable to the graph under test. Disabling it makes the two passes structurally identical. 
This also avoids the separate question of whether a model's cached path agrees with its full-sequence path, which is a different property (§3.3, B6).

\textbf{(4) Deterministic single-precision evaluation.} In a correct implementation $\Delta$\_$\ell$ is not "small,
" it is \textbf{exactly zero} at every layer (§3.4) --- the two computations perform bitwise identical arithmetic on the compared positions. 
This is what makes the verdict threshold-robust: with clean deltas at exact zero, any $\tau \in [10^{-6},10^{-3}]$ returns the same answer on our test set.
Precision does matter for the \textit{magnitude of leak} the test can localize, which we characterize as an operating range in §3.5 rather than treating as a fixed limit.

\subsection{Cost}
Two forward passes, no backward pass, no gradient graph, no optimizer state, no training data, no labels, and no accelerator. 
Memory overhead is \textit{L} captured activations of shape \textit{T}$\times$\textit{d}; at \textit{T} = 48 this is negligible relative to the weights. 
In our runs a full scan of a 135M-parameter model on CPU took \textbf{0.1--0.5 s} after loading. 
The census and injection suite ran on CPU; the multi-billion-parameter audits (§3.7.4, §3.8) used a single GPU for capacity, not for the test.

Our objective is complementary to efficiency-oriented advances such as FlashAttention
\citep{dao2022flashattention}.
Whereas such methods improve throughput and memory efficiency,
our focus is verification of causal correctness.

\section{Experiments}
\begin{figure}[t]
\centering
\includegraphics[width=0.82\linewidth]{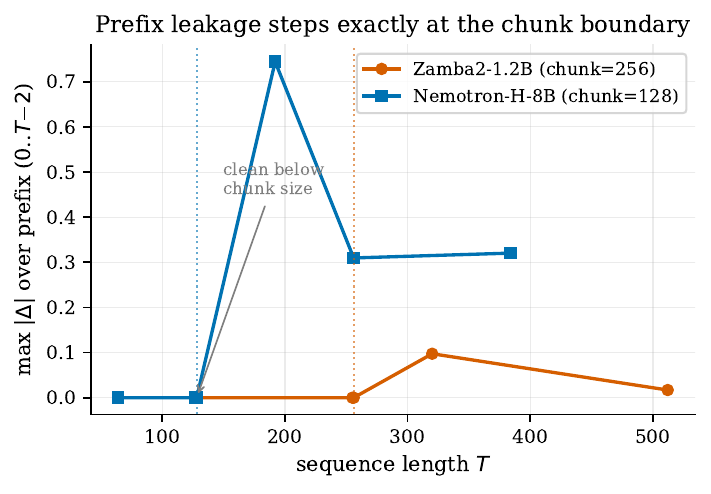}
\caption{Prefix leakage begins exactly at each model's chunk boundary (Zamba2 at 256, Nemotron-H at 128).}
\label{fig:chunk}
\end{figure}
\begin{figure}[t]
\centering
\includegraphics[width=0.82\linewidth]{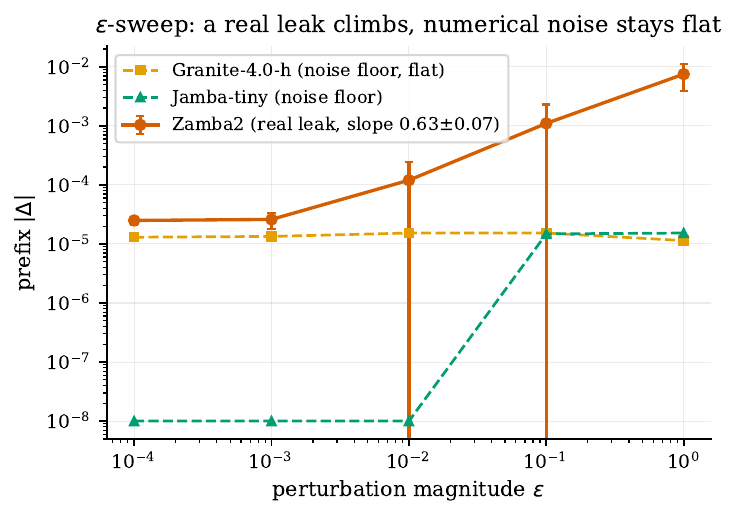}
\caption{The $\varepsilon$-sweep discriminator: a real leak climbs with perturbation magnitude (Zamba2, slope $0.63\pm0.07$) while numerical floors stay flat.}
\label{fig:eps}
\end{figure}
\begin{figure}[t]
\centering
\includegraphics[width=0.82\linewidth]{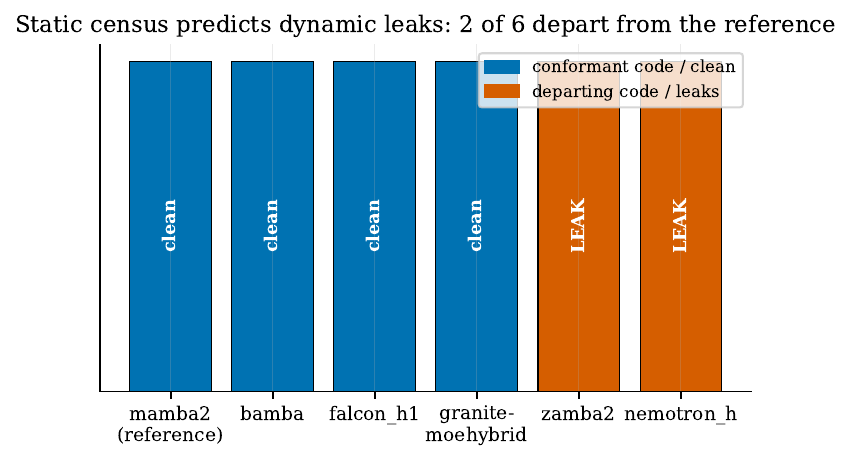}
\caption{Static code census predicts dynamic leaks: two of six chunked-scan implementations depart from the reference and both leak.}
\label{fig:census}
\end{figure}
\begin{figure}[t]
\centering
\includegraphics[width=0.82\linewidth]{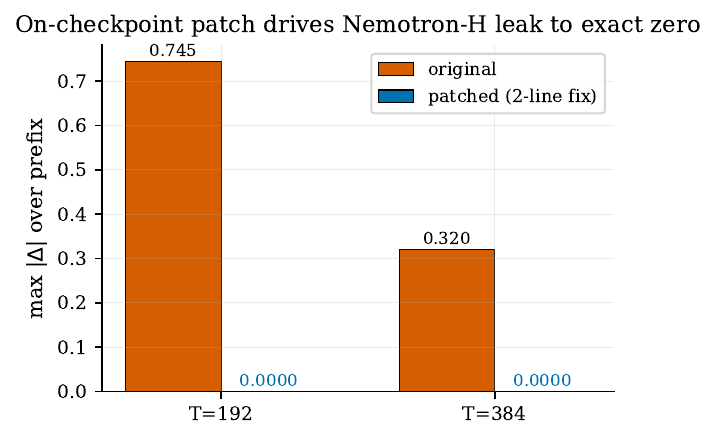}
\caption{The two-line fix drives the Nemotron-H prefix leak to exact zero on the released checkpoint.}
\label{fig:patch}
\end{figure}

\subsection{Setup}

The census and injected-fault suite ran on CPU (Intel Xeon Gold 6526Y, 64 threads, 251 GB RAM); 
\textbf{the test itself needs no accelerator.} The larger checkpoints --- our own 7B model (§3.7.4) 
and the 1.2B--8B public models of §3.8 --- were audited on a single GPU for memory capacity only, not because the method requires one. 
Unless stated otherwise: float32, \textit{T} = 48, $\tau$ = 1e-6, fixed seed, caching disabled, models loaded via a standard model hub in evaluation mode.

Two harnesses were used. \texttt{leak\_cpu\_suite2.py} performs the census with baselines B1--B3; 
\texttt{baseline\_e2.py} performs the extended six-baseline comparison. They implement the same detector and agree on all shared models (§3.3).

\textbf{Injection protocol (positive control).}
For each model and each of the 8 patterns in §3.2,
we wrap one layer's forward function to apply the leak to
that layer's output, at three depths---an early layer
(index 1), a middle layer ($\lfloor L/2 \rfloor$),
and a late layer ($L-2$)---giving \textbf{24 trials per model}.
A trial counts as \texttt{exactly localized} only if the
reported first offending layer equals the injected layer.
Injection strength is $\varepsilon = 1.0$ unless a sweep is stated.

\begin{quote}
\textbf{Reporting rule adopted after §3.7.} A CLEAN verdict for a checkpoint is reported only if that checkpoint's own 24 positive-control trials passed. 
Checkpoints whose positive controls failed are reported separately as \textit{not audited}, never as clean.
\end{quote}

\subsection{Positive control: injected faults}

Across \textbf{8 public checkpoints $\times$ 8 patterns $\times$ 3 depths = 192 trials}, the test localized the injected fault to the exact layer in \textbf{192/192} cases. 
Per-checkpoint results appear in the census table (§3.6, Table 4); no checkpoint scored below 24/24.

This holds uniformly across the four taxonomy classes and across five architecture families, including two with no attention mask anywhere in the stack (\texttt{mamba-130m-hf}, 
\texttt{rwkv-6-world-1b6}). That uniformity is the point of an output-based test: it never asks what kind of mixer produced the representation.

\subsection{Comparison against existing practice}

We implemented six alternative auditing practices and evaluated them on the same injected faults across four architectures 
(\texttt{SmolLM2-135M}/Llama, \texttt{Qwen3-0.6B}/Qwen3, \texttt{pythia-1.4b}/GPTNeoX, and \texttt{mamba-130m-hf}/Mamba), 
yielding a total of \textbf{96 trials}. As shown in Table~\ref{tab:baseline-comparison}, our method detected and correctly localized all 96 injected faults, 
whereas existing lightweight baselines either failed to detect certain leakage patterns or provided no localization information.

\begin{table}[t]
\centering
\small
\setlength{\tabcolsep}{4pt}

\caption{Baseline comparison on 96 injected faults across four architectures. 
Our method and the gradient-based baseline (B5) both achieve perfect localization, 
whereas logits-only perturbation (B1) detects leaks but provides no localization. 
Static attention-mask inspection (B2) detects none of the injected faults.}
\label{tab:baseline-comparison}

\begin{tabular}{
p{0.35\linewidth}
c
c
c
c
p{0.18\linewidth}
}
\toprule
Method &
Fwd &
Bwd &
Detected &
\textbf{Exactly localized} &
False positive on clean model \\
\midrule

\textbf{Ours} (per-layer $\Delta$)
& 2 & 0 & 96/96 & \textbf{96/96}
& none ($\Delta=0$) \\

B1 --- logits-only perturbation
& 2 & 0 & \textbf{96/96}
& 0/96
& none \\

B2 --- static attention-mask inspection
& 0 & 0 & \textbf{0/96}
& 0/96
& none \\

B3 --- shuffled-suffix perplexity
& 2 & 0
& 71/96
& 0/96
& none \\

B4 --- future-token resampling, 3 seeds
& 4 & 0
& \textbf{96/96}
& 0/96
& none \\

B5 --- gradient of prefix outputs w.r.t.\ future positions
& 1
& \textbf{1}
& \textbf{96/96}
& \textbf{96/96}
& none \\

B6 --- cached vs.\ full-sequence consistency
& T+1=17
& 0
& 24/24$\dagger$
& 0/24
& \textbf{yes: $\Delta=1.850\times10^{-4}$ on a clean model} \\

\bottomrule
\end{tabular}
\end{table}

† B6 was measured on \texttt{SmolLM2-135M} only; its cost is linear in sequence length.

Four readings, in decreasing order of comfort.

\textbf{(a) Mask inspection sees nothing --- 0/96.} This is stronger than we hypothesized. We expected the four aggregation- 
and direction-class patterns to be invisible to mask inspection; in fact \textbf{all eight} are, including the off-by-one class. 
The reason is structural: all eight faults live in a layer's output, so the mask attribute remains correctly set. 
On \texttt{SmolLM2-135M} the inspector found 30 modules exposing a causal attribute and flagged none of them, on every one of the 24 injected runs. 
On \texttt{mamba-130m-hf} the procedure has no input at all: zero modules expose such an attribute.

\textbf{(b) Shuffled-suffix perplexity fails structurally on the most common bug class.} B3 misses 25 of 96, and the misses are not random. 
In every case where B3 misses a C1 fault, the loss difference is \textbf{exactly 0.000e+00} --- not small, zero --- so no choice of threshold recovers it. 
The reason is a mismatch of radius: B3 shuffles positions $\geq$ cut and evaluates targets < cut, 
but a radius-$1$ leak reaches at most position $cut-1$ and therefore never touches the shuffled region.
\textbf{B3 is structurally blind to local leaks whose radius is shorter than the distance to the shuffle boundary --- 
which is precisely the off-by-one class.} Separately, one C3 fault on Qwen3 produced a loss difference of 5.63e-05, 
below B3's 1e-4 threshold: a threshold miss rather than a structural one. We distinguish the two, because only the first is unfixable.

\textbf{(c) We do not improve on existing practice at detection.} B1 and B4 detect every injected fault, exactly as our method does. 
Any claim that current practice fails to \textit{notice} leakage is not supported by our measurements. What B1 and B4 cannot do is say where: 
\texttt{located\_layer} is \texttt{None} in 96/96 trials, and not for want of implementation effort --- the information is not present in the output logits. \textbf{Our contribution is localization.}

\textbf{(d) Gradient-based checking matches our localization exactly.} B5 achieves 96/96 detection and 96/96 exact localization --- identical to ours. 
We report this prominently rather than burying it. Our advantage over B5 is cost and robustness of applicability, not accuracy:

\begin{itemize}
\item B5 requires a \textbf{backward pass} and retention of the full activation graph; ours runs entirely under no-grad with two forwards.
\item B5 requires every operation on the path to be differentiable and to have a working gradient implementation --- which excludes quantized weights, several custom kernels, and non-differentiable ops.
\item A zero gradient does not imply absence of dependence (a saturated nonlinearity has zero gradient at a point while still transmitting information), 
so B5's negative results carry a caveat that a finite-difference comparison does not.
\end{itemize}

We therefore position B5 as a \textbf{valid alternative method with the same localization power and a higher cost profile}, not as a weaker baseline.

\textbf{(e) Cache-consistency checking raises false alarms on correct models.} B6 detected all 24 injected faults on \texttt{SmolLM2-135M}, 
but also fired on the \textbf{clean} model with max|$\Delta$| = 1.850e-04 against its 1e-4 threshold. 
Injected faults produce $\Delta$ of order 10\textasciicircum{}1, so the classes are separable in principle, 
but the threshold must be recalibrated per model and per precision. The contrast with our method is the sharpest argument in this table: 
\textbf{our clean deltas are exactly zero, so no calibration is required.} We note that B6 measures a genuinely different property (agreement between the cached and uncached code paths), 
which is worth checking for its own reasons.

\subsection{Negative control: clean deltas are exactly zero}

Table~\ref{tab:bit-level-verification} reports the bit-level verification results on four clean checkpoints. Across all inspected layers, 
repeated evaluations produced identical outputs under tensor equality, differing-element counts, 
and bit-pattern comparison. No differing elements were observed, and the maximum absolute deviation was exactly zero for every model.

\begin{table}[t]
\centering
\small

\caption{Bit-level verification on clean checkpoints. For each checkpoint, all compared layer outputs are exactly identical under repeated evaluation, with zero differing elements and zero maximum absolute difference.}
\label{tab:bit-level-verification}

\begin{tabular}{lccccc}
\toprule
\makecell{Model\\checkpoint} &
\makecell{Layers\\checked} &
\makecell{All layers\\\texttt{torch.equal}} &
\makecell{Differing\\elements} &
\makecell{max\\abs} &
\makecell{Bit patterns\\identical} \\
\midrule

HuggingFaceTB/SmolLM2-135M &
30 &
[OK] &
\textbf{0} &
0.0 &
[OK] \\

Qwen/Qwen3-0.6B &
28 &
[OK] &
\textbf{0} &
0.0 &
[OK] \\

EleutherAI/pythia-1.4b &
24 &
[OK] &
\textbf{0} &
0.0 &
[OK] \\

state-spaces/mamba-130m-hf &
24 &
[OK] &
\textbf{0} &
0.0 &
[OK] \\

\bottomrule
\end{tabular}

\end{table}

This exactness has two implications. First, the method is threshold-robust: clean deltas remain at zero, while injected faults produce substantially larger values, 
so any $\tau \in [10^{-6}, 10^{-3}]$ yields the same verdict on our test set. Second, causal correctness is a property of the computation graph rather than the specific weights. 
The result follows from identical computations on the same prefix and remains consistent across checkpoints of different architectures and provenance.

\subsection{Operating range: precision and threshold}

The test's ability to identify the \emph{exact} offending layer depends on the magnitude of the injected leak relative to numerical precision 
and the decision threshold. Rather than treating this as a fixed limit, we characterize it as an \textbf{operating range}. 
Table~\ref{tab:localization-floor} reports the smallest injection strength $\varepsilon$ that can still be localized to the correct layer as a function of arithmetic precision, 
threshold $\tau$, and injection depth. 
The experiment uses \texttt{SmolLM2-135M} with the C2 \texttt{pool\_leak} fault, sweeping $\varepsilon$ over 15 orders of magnitude at three injection depths (layers 1, 15, and 28).

\begin{table}[t]
\centering
\small

\caption{Exact-localization floor: smallest $\varepsilon$ still localized to the exact injected layer. Lower values indicate higher sensitivity of the audit to weak temporal-leakage signals.}
\label{tab:localization-floor}

\begin{tabular}{lll}
\toprule
Precision & $\tau = 10^{-6}$ & $\tau = 10^{-14}$ \\
\midrule

float32 &
$10^{-6}$ (L1, L15); $10^{-7}$ (L28) &
\textbf{$10^{-10}$} (all three depths) \\

float64 &
$10^{-5}$ (all three depths) &
\textbf{$10^{-13}$} (L1, L15);
\textbf{$10^{-14}$} (L28) \\

\bottomrule
\end{tabular}

\end{table}

Three observations, all of which we consider necessary for honest reporting.

\textbf{(1) At the conventional threshold, $\tau$ --- not precision --- sets the floor.} At $\tau$ = 1e-6 the float64 floor (1e-5) looks \textit{worse} than float32 (1e-6). This is an artifact, 
not a property. In float64 the measured delta is exactly proportional to $\varepsilon$: at layer 15 the ratio $\Delta$/$\varepsilon$ is 0.6791552 
and stays constant to seven significant figures from $\varepsilon$ = 1e-1 down to $\varepsilon$ = 1e-11. In float32 that proportionality breaks below $\varepsilon$ $\approx$ 1e-6, 
and the deltas that cross $\tau$ at $\varepsilon$ = 1e-9 (e.g. 1.49e-08 at layer 15, a power-of-two value) are \textbf{rounding residue, not signal}. 
Part of float32's apparent sensitivity at the conventional threshold is numerical noise that happens to exceed the threshold.

\textbf{(2) Because clean deltas are exactly zero, $\tau$ can be lowered without incurring false positives.} This is what converts the exact-zero property from a curiosity into a usable lever: 
at $\tau$ = 1e-14 the floors improve to 1e-10 (float32) and 1e-13--1e-14 (float64) --- three to four orders of magnitude.

\textbf{(3) float32 has a hard floor that no threshold recovers.} At $\varepsilon$ $\leq$ 1e-11 the float32 delta at the injected layer is \textbf{exactly 0.0}: 
the perturbation has fallen below the representable resolution of the activations, so the leak is not merely undetected, it is absent from the computed tensor. 
float64 still carries nonzero deltas at $\varepsilon$ = 1e-15 (1.78e-15 to 7.11e-15). \textbf{Auditing for leaks weaker than \textasciitilde{}1e-10 relative magnitude requires float64.}

Below the exact-localization floor the test does not fail silently: it continues to report a leak, but names a layer one to a few positions downstream, 
because the sub-resolution signal at the true layer is amplified by subsequent layers before it crosses $\tau$. Detection degrades later than localization.

\textbf{Per-checkpoint floors vary and should be measured, not assumed.} At $\tau$ = 1e-6 with C2 injection at the middle layer, 
the exact-localization floor was 1e-5 for \texttt{Qwen3-0.6B}, \texttt{mamba-130m-hf}, \texttt{rwkv-6-world-1b6}, 
and \texttt{recurrentgemma-2b}; 1e-7 for \texttt{gemma-3-1b-it}; and for \texttt{LFM2-1.2B} no sweep point at $\varepsilon$ $\leq$ 1e-1 was exactly localized (injection 
at layer 8 was consistently reported at layer 11), although all 24 of its $\varepsilon$ = 1.0 trials were exact. We have not determined the cause of the \texttt{LFM2-1.2B} outlier 
and do not draw a conclusion from it; we report it because omitting it would misrepresent the floor as model-independent.

\subsection{Census of public checkpoints}

To reduce the risk of self-evaluation bias, we applied the audit unchanged to a diverse set of public checkpoints spanning attention, state-space, recurrent, 
and hybrid architectures. Table~\ref{tab:checkpoint-census} summarizes the census results. 
All eight checkpoints produced identical outcomes on the core audit metrics: \texttt{clean max = 0.000e+00}, \texttt{Verdict = CLEAN}, \texttt{Ours (exact) = 24/24}, and \texttt{B1 = 24}. 
Because these values were invariant across the entire set, they are omitted from the table for brevity. 
The census therefore supports the claim that the audit applies uniformly across substantially different mixer families.

\begin{table}[t]
\centering
\small

\caption{Public checkpoint census conducted in float32 on CPU with $\tau = 10^{-6}$ and sequence length $T = 48$. 
The table summarizes the eight checkpoints audited in the census, together with architecture class, model size, 
and the outcomes of baselines B2 (static attention-mask inspection) and B3 (shuffled-suffix perplexity).}
\label{tab:checkpoint-census}

\begin{tabular}{lllllll}
\toprule
\# & Checkpoint & Mixer class & L & Params & B2 & B3 \\
\midrule

1 & HuggingFaceTB/SmolLM2-135M & attention & 30 & 134M & 0 & 18 \\
2 & Qwen/Qwen3-0.6B & attention & 28 & 596M & 0 & 17 \\
3 & EleutherAI/pythia-1.4b & attention & 24 & 1.41B & 0 & 18 \\
4 & google/gemma-3-1b-it & attention & 26 & 1.00B & 0 & 18 \\
5 & state-spaces/mamba-130m-hf & state-space (no mask) & 24 & 129M & 0 & 18 \\
6 & RWKV/rwkv-6-world-1b6 & linear recurrent (no mask) & 24 & 1.60B & 0 & 17 \\
7 & google/recurrentgemma-2b & recurrent + attention & 26 & 2.68B & 0 & 18 \\
8 & LiquidAI/LFM2-1.2B & conv + attention hybrid $\dagger$ & 16 & 1.17B & 0 & 17 \\

\midrule

& \textbf{Total} & {} & {} & {} &
\textbf{0} & 141 \\

\bottomrule
\end{tabular}

\end{table}

† \texttt{LFM2-1.2B}'s configuration declares per-layer types explicitly: convolution at 10 of 16 layers and full attention at layers 2, 5, 8, 10, 12, 14.

Three conclusions.

\textbf{(a) The method applies across mixer families without modification.} The same code produced verdicts for masked attention, unmasked state-space scans, 
linear recurrences, and convolution/attention hybrids. Two of the eight checkpoints (\#5, \#6) have no attention mask anywhere, so the incumbent audit procedure has no input for them at all.

\textbf{(b) Exact-zero deltas reproduce on models we did not write.} This rules out the possibility that exactness is an artifact of our own implementation conventions.

\textbf{(c) The test does not raise false alarms.} All eight checkpoints are clean while the same detector, on the same checkpoints, localized 192 injected faults. 
A detector that flagged models indiscriminately would not show this pattern.

\textbf{What this table does and does not establish.} Every checkpoint in Table 4 is clean \textit{at the test length used here}, T = 48. 
The claim supported by these results is that the test applies to any architecture and does not fire spuriously. 
It is not evidence that released models are causally correct in general --- as §3.8 shows, this very census was run below the sequence length at which one class of defect becomes observable at all.

\subsection{When the audit itself fails}

This section reports a negative result about our own procedure. It changed the paper's methodology, and it is the finding we would most want a reader to take away.

\subsubsection{Three checkpoints that looked clean and were not audited}

Three checkpoints of one hybrid family --- \texttt{tiiuae/Falcon-H1-0.5B-Base} (36 layers, 521,411,104 params), \texttt{Falcon-H1-1.5B-Instruct} (24 layers, 1,554,863,488), 
and \texttt{Falcon-H1-7B-Base} (44 layers, 7,585,648,736) --- returned \textbf{max|$\Delta$| = 0.0 at every layer}. Read naively, that is three clean verdicts, and they would have entered Table 4.

Their positive controls returned \textbf{0/24}. Every injected fault, at every depth, including the largest-magnitude patterns at $\varepsilon$ = 1.0, 
produced $\Delta$ = 0.0 at the injected layer --- identical to the clean scan. The sensitivity sweep likewise reported no detection at any $\varepsilon$ from 1e-1 to 1e-9.

Diagnosis. We verified that our injection wrapper was invoked (the wrapped forward executed and returned a tuple whose first element is a rank-3 tensor), 
so the instrumentation was attached correctly. We then compared the two inputs at the output:

\begin{itemize}
\item Full logit difference \textbf{including} the differing final position: \textbf{0.0}
\item Difference at the final position alone: \textbf{0.0}
\item Difference between the token embeddings of the two differing tokens: \textbf{3.78e-23}
\end{itemize}

The models, as loaded in our environment, \textbf{produce bit-identical outputs for different inputs} --- they do not respond to their input at all, 
while still producing plausibly-scaled, position-varying logits (|logit| up to 45.4, no NaNs). Under such a load, $\Delta$ = 0 everywhere is guaranteed regardless of whether the architecture is causal, 
and carries no information.

Our environment emitted two relevant warnings during load: the optional fused sequence-model kernels were unavailable (falling back to a reference implementation), 
and compiled extensions were skipped as incompatible with the installed framework version. We did not isolate which of these, 
if either, causes the behavior, and we make \textbf{no claim about these models' correctness}. 
The correct statement is: \textit{in our environment these checkpoints did not load into a state where they could be audited.}

\subsubsection{The gate this implies}

\begin{quote}
\textbf{A clean verdict is admissible only if the positive control passes on the same loaded checkpoint.}
\end{quote}

The check is nearly free --- the same 24 injections already used for validation --- and without it this paper would have published three false clean verdicts on models whose loaded state was degenerate. 
We know of no reason this failure mode is specific to us: \textbf{any} audit that concludes from an absence of signal is vulnerable to a silently inert system under test, 
and audits reporting negative results should be expected to demonstrate that their instrument was live on that specific artifact.

This is why Table 4 reports "Ours (exact)" beside every verdict rather than in a separate validation section. The two numbers are one claim.

\subsubsection{Checkpoints we could not audit at all}

To avoid survivorship bias, we report every checkpoint that was attempted but did not yield an audit verdict rather than silently excluding failed runs. 
Table~\ref{tab:attempted-not-audited} summarizes these cases and the reason each audit could not be completed. 
The causes include environment and loading failures, unavailable custom CUDA dependencies, packaging defects, and repository-access restrictions. 
Reporting these outcomes is important because omission would artificially inflate the apparent audit coverage and obscure practical barriers to reproducibility.

\begin{table}[t]
\centering
\small

\caption{Attempted but not audited. These checkpoints were examined during the census but could not be evaluated because of loading failures, missing dependencies, 
packaging issues, or repository-access restrictions.}
\label{tab:attempted-not-audited}

\begin{tabular}{p{0.28\linewidth} p{0.42\linewidth} p{0.18\linewidth}}
\toprule
Checkpoint & Reason & Nature \\
\midrule

tiiuae/Falcon-H1-0.5B / 1.5B / 7B &
Loaded model does not respond to input; positive control 0/24 &
Load/environment failure (§3.7.1) \\

nvidia/Hymba-1.5B-Base &
Requires \texttt{causal\_conv1d}, \texttt{mamba\_ssm} &
Custom CUDA kernel dependency \\

microsoft/Phi-4-mini-flash-reasoning &
Requires \texttt{causal\_conv1d}, \texttt{causal\_conv1d\_cuda}, \texttt{mamba\_ssm} &
Custom CUDA kernel dependency \\

togethercomputer/StripedHyena-Nous-7B &
Requires a fused normalization kernel from a source build &
Custom CUDA kernel dependency \\

state-spaces/mamba2-2.7b &
No \texttt{model\_type} in config; not loadable through the standard interface &
Packaging \\

Zyphra/Zamba2-7B &
Gated repository; access not granted &
Access \\

\bottomrule
\end{tabular}

\end{table}

The three kernel-dependency rows are worth stating as a finding rather than an inconvenience: \textbf{a portion of the hybrid model ecosystem cannot be independently audited on commodity hardware}, 
because loading the model at all requires GPU-only compiled kernels. Any conformance-certification norm has to contend with that, 
and it argues for certificates produced by the releasing party, who can run the model, rather than by third parties who often cannot.

\subsubsection{Our own released models: the load failure, its cause, and the completed audit}

An earlier draft of this section reported that four of our own publicly released checkpoints \textbf{failed to load} with a device-placement error (\texttt{Tensor on device meta is not on the expected device cpu}), 
and that completing their audit was a prerequisite for submission rather than future work. We have since diagnosed the failure and completed the audit for the primary release.

\textbf{Cause.} The failure is not in the checkpoint but in the loading path. 
\texttt{from\_pretrained} instantiates parameters on the \texttt{meta} device before materializing them, 
while this architecture's rotary cache is constructed on an explicitly hard-coded CPU device at \texttt{\_\_init\_\_} time. The two are incompatible, 
and --- importantly --- passing \texttt{low\_cpu\_mem\_usage=False} does \textbf{not} avoid it, because the meta-device stage is internal to that path.

\textbf{Resolution.} Instantiating from the configuration and loading weights separately bypasses the conflict entirely:

\begin{verbatim}
cfg = AutoConfig.from_pretrained(rid, trust_remote_code=True)
torch.set_default_dtype(torch.bfloat16)
m = AutoModelForCausalLM.from_config(cfg, trust_remote_code=True) 
    # real device, no meta stage
torch.set_default_dtype(torch.float32)
m.load_state_dict(load_file(hf_hub_download(rid, "model.safetensors")),
     strict=False)
m = m.to("cuda").eval()
\end{verbatim}

This loads with 0 missing and 0 unexpected tensors and generates coherently. 
We report the diagnosis because the same conflict will affect any architecture that builds device-pinned buffers during \texttt{\_\_init\_\_}, 
and because a model that cannot be loaded cannot be audited by anyone --- the point made in §3.7.3.

\textbf{Audited subject.} \texttt{FINAL-Bench/Aether-7B-5Attn} --- 49 decoder layers composed of seven mixer types with exactly seven layers each (NSA, differential, full, linear, 
sliding-window, compressive, and a gated hybrid), arranged so that every window of seven consecutive layers contains each type exactly once. 
The architecture, its training, and an ablation separating placement from composition are the subject of a \textbf{companion paper}; here we treat it purely as an audit target.

Table~\ref{tab:self-audit} shows that our released checkpoint remains clean under negative controls and correctly localizes all injected positive-control faults.

\begin{table}[t]
\centering
\small

\caption{Audit of our own released checkpoint (float32, $T = 48$, $\tau = 10^{-6}$). The checkpoint passes both negative and positive controls, with exact localization of all injected faults and no leakage observed under normal operation.}
\label{tab:self-audit}

\begin{tabular}{ll}
\toprule
Check & Result \\
\midrule

Negative control (49 layers) &
max $\Delta = 0.000\mathrm{e}{+00}$ \\

Negative control, after loading trained weights &
unchanged $0$ \\

\textbf{Positive control (4 sites $\times$ 4 fault classes)} &
\textbf{16/16 exact localization} \\

\bottomrule
\end{tabular}

\end{table}

Positive-control sites were chosen to span mixer families: L0 (NSA), L12 (gated hybrid), L25 (NSA), L40 (linear attention). 
Every injection was localized to the exact layer. This is a reduced grid --- 4 sites $\times$ 4 fault classes = 16 trials --- 
rather than the 8 patterns $\times$ 3 depths = 24 used on the small-model census (§3.2). The reduction is deliberate: 
on a 7B model the injection sites are pinned to the \textit{distinct mixer families actually present in this stack}, 
since the goal here is to certify that each such family is live and localizable on the loaded checkpoint, not to re-run the generic fault taxonomy already validated 192/192 at small scale.

We ran the positive control on the same loaded checkpoint precisely because §3.7.1 showed that a clean verdict without a liveness demonstration is worthless. 
Applying that standard to our own release rather than only to third-party models is the point of this subsection.

\textbf{What remains.} The instruction-tuned variant \texttt{Aether-7B-5Attn-it} passed the negative control on all 49 layers but has not yet received a positive control. 
Two further internal checkpoints remain unaudited. We do not claim our stack is causally correct beyond what Table 6 reports.

\subsection{A causal defect in a released model}

Everything reported up to this point was measured at a single short test length, T = 48. That choice was never justified; it was inherited. 
This section reports what happened when we questioned it.

\subsubsection{Why the test length matters}

A hybrid stack mixes operators whose behavior is governed by internal size parameters --- a chunk length for chunked state-space scans, a window width for local attention, 
a kernel width for causal convolution. \textbf{If the test sequence is shorter than such a parameter, the corresponding code path is never exercised in its general form.} 
A chunked scan over a sequence shorter than one chunk degenerates to a single chunk with no inter-chunk recurrence at all; a sliding window wider than the sequence degenerates to full attention.

An audit conducted below those thresholds therefore cannot observe a whole class of defects, 
and will report the model as clean. We re-ran the audit with sequences chosen to exceed each model's declared chunk or window parameter, 
and extended the census by twelve further checkpoints, including three at 7B--9B scale (Nemotron-H-8B, Bamba-9B, falcon-mamba-7B).

\subsubsection{A code-level census, and the prediction it made}

Rather than audit models blindly, we first examined the source code. The inter-chunk recurrence of a Mamba-2-style scan is a compact and recognizable computation, 
and the reference implementation (\texttt{modeling\_mamba2.py}) defines a specific orientation of the input- and output-chunk axes. 
We therefore performed a static census of every chunked-scan implementation shipped with \texttt{transformers}~5.7.0 before running any model. 
As summarized in Table~\ref{tab:static-census}, four implementations matched the reference, while \texttt{Zamba2} and \texttt{Nemotron-H} exhibited the same axis-handling deviation. 
This static analysis required neither model weights nor forward execution and served as a blind prediction later tested dynamically.

\begin{table}[t]
\centering
\small

\caption{Static census of chunked-scan implementations (\texttt{transformers} 5.7.0). 
Two implementations (\texttt{modeling\_zamba2.py} and \texttt{modeling\_nemotron\_h.py}) depart from the reference inter-chunk recurrence 
by reducing over the output-chunk axis rather than the input-chunk axis.}
\label{tab:static-census}

\begin{tabular}{lll}
\toprule
Implementation & Inter-chunk axis handling & Class \\
\midrule

\texttt{modeling\_mamba2.py} (reference) &
\texttt{transpose(1,3)} $\rightarrow$ reduce over \textbf{input} chunk &
reference \\

\texttt{modeling\_bamba.py} &
matches reference &
[OK] conformant \\

\texttt{modeling\_falcon\_h1.py} &
matches reference &
[OK] conformant \\

\texttt{modeling\_granitemoehybrid.py} &
matches reference &
[OK] conformant \\

\textbf{\texttt{modeling\_zamba2.py}} &
\texttt{permute} $\rightarrow$ reduce over \textbf{output} chunk &
\textbf{departs} \\

\textbf{\texttt{modeling\_nemotron\_h.py}} &
\texttt{permute} $\rightarrow$ reduce over \textbf{output} chunk &
\textbf{departs} \\

\bottomrule
\end{tabular}

\end{table}

The static census produced a concrete and falsifiable prediction. As shown in Table~\ref{tab:dynamic-audit}, 
two of the six inspected implementations were flagged as suspect, while the remaining four were classified as conformant. Importantly, the \texttt{Zamba2} 
and \texttt{Nemotron-H} inter-chunk recurrence blocks are byte-for-byte identical (§3.8.4), 
indicating a shared code lineage rather than two independent implementation errors. 
For \texttt{Nemotron-H}, this constituted a genuine blind prediction because we had not examined the model before performing the census. For \texttt{Zamba2}, 
the census independently re-derived, from source code alone, a defect that we had previously discovered and reported. 
Table~\ref{tab:dynamic-audit} tests this prediction by auditing each model at sequence lengths that exceed its declared chunk or window parameter.

\begin{table}[t]
\centering
\small

\caption{Dynamic audit at chunk-exceeding sequence lengths (testing the prediction). Models identified by the static census 
as deviating from the reference implementation (\texttt{Zamba2} and \texttt{Nemotron-H}) exhibit causal leakage precisely at their declared chunk boundaries, 
whereas all conformant implementations remain clean.}
\label{tab:dynamic-audit}

\begin{tabular}{lll}
\toprule
Checkpoint & Declared parameter & Verdict \\
\midrule

\textbf{Zamba2-1.2B} &
\texttt{chunk\_size} 256 &
\textbf{leak, onset = 256} \\

\textbf{Nemotron-H-8B} &
\texttt{chunk\_size} 128 &
\textbf{leak, onset = 128} \\

Bamba-9B &
\texttt{mamba\_chunk\_size} 256 &
clean \\

Falcon-Mamba-7B &
\texttt{conv\_kernel} 4 &
clean \\

Falcon-H1-1.5B &
\texttt{mamba\_chunk\_size} 128 &
clean \\

Granite-4.0-H-Tiny &
\texttt{mamba\_chunk\_size} 256 &
clean (see §3.8.6) \\

Gemma-3-1B &
\texttt{sliding\_window} 512 &
clean at $T = 1536$ \\

RecurrentGemma-2B &
\texttt{sliding\_window} 2048 &
clean at $T = 3072$ \\

Gemma-2-2B &
\texttt{sliding\_window} 4096 &
clean \\

Mamba2-130M, 370M &
\texttt{chunk\_size} 256 &
clean \\

RWKV-6-1.6B, Qwen3-0.6B,\\
LFM2-1.2B, ZR1-1.5B &
--- &
clean \\

\bottomrule
\end{tabular}

\end{table}

\textbf{The prediction holds.} Both departing implementations leak, each at its own declared chunk boundary (256 for Zamba2, 128 for Nemotron-H); 
no conformant implementation leaks under audit. Scale confers no protection: an \textbf{8B} model (Nemotron-H) leaks while a \textbf{9B} model (Bamba) does not.

We are careful about the two strengths of the conformant-clean half of the claim, because the positive-control gate of §3.2 must apply here too. \textbf{At the code level it is firm.} 
A reference \texttt{Mamba2} built from a random configuration is \textit{gated}-clean: exact-zero prefix deltas at every length up to several multiples of its chunk size, 
\textbf{and} a live positive control that localizes injected faults to their origin layer even as the injected signal propagates to every layer downstream (§3.8.4). 
That rules out the inert-load false-clean of §3.7. \textbf{At the level of individual released checkpoints it is weaker, and we say so.} 
We audited every conformant checkpoint and found no leak, but the injection gate could not be applied uniformly --- some architectures' layer return signatures do not accept our fault injection, 
so their positive control cannot be run in our harness --- 
and so those particular clean verdicts are \textbf{provisional} (no live-detector guarantee) rather than gated (§4). 
No \textit{gated} checkpoint contradicts the prediction. Two clean results are additionally reassuring on their own terms: 
\texttt{gemma-3-1b-it} and \texttt{recurrentgemma-2b} were audited at three times their own window width and still returned exact-zero deltas, 
so the positive findings are not an artifact of merely lengthening sequences.

\textbf{Which execution path this is --- and is not.} The defect lives in the chunked-scan written in PyTorch (\texttt{torch\_forward} / \texttt{segment\_sum}). 
This path executes whenever the optional fused kernels (\texttt{mamba\_ssm}, \texttt{causal\_conv1d}) are \textbf{absent} --- which is the case for all CPU execution, 
for any GPU environment without those packages installed, and for a stock \texttt{transformers} install that has not added them (they are separate, 
frequently uninstallable dependencies; in our environment they refuse to build against the current torch). It is the path exercised by a large fraction of evaluation, research, 
and reproduction workflows. It is \textbf{not} necessarily the path a production server takes: an install with the fused CUDA kernels present dispatches to a different implementation, 
which we did not audit because those kernels do not build in our environment. We therefore scope the claim precisely: 
\textbf{the reference PyTorch implementation shipped in \texttt{transformers} for these two models violates causality across a chunk boundary; 
whether their fused-kernel path shares the defect is untested and open.} The finding matters because the unaudited-fast-path caveat cuts the other way too --- 
a model can pass every fused-kernel test and still produce leaking logits the moment it is run without the kernels, e.g. on CPU or in CI.

\subsubsection{The defect}

For \texttt{Zyphra/Zamba2-1.2B}, perturbing a single token at position \textit{P} = 400 in a sequence of length 512 changes the logits at earlier positions:

\begin{verbatim}
pos   0 : 0.0000e+00
pos 100 : 0.0000e+00
pos 200 : 0.0000e+00
pos 255 : 0.0000e+00     <- clean up to here
pos 256 : 5.7473e-03     <- contamination begins
pos 300 : 5.0647e-03
pos 399 : 3.2187e-03
\end{verbatim}

Contamination begins at position \texttt{256}, 
which is exactly the model's declared \texttt{chunk\_size}. 
The measured value is $\max |\Delta(0..399)| = 1.1241\times10^{-2}$.

\texttt{nvidia/Nemotron-H-8B-Base-8K} behaves identically at its own boundary. 
Perturbing only the final token and sweeping the sequence length gives a clean-then-contaminated step exactly at \texttt{chunk\_size} = 128:

\begin{quote}
   \ttfamily T = 64 \hspace{1em}: max $|\Delta(\mathrm{prefix})|$ = 0.0000e+00 \hfill <- one chunk; clean \\ 
   T = 127 \hspace{1em}: max $|\Delta(\mathrm{prefix})|$ = 0.0000e+00 \\
   T = 128 \hspace{1em}: max $|\Delta(\mathrm{prefix})|$ = 0.0000e+00 \hfill <- clean up to the \\ 
     boundary \\
   T = 192 \hspace{1em}: max $|\Delta(\mathrm{prefix})|$ = 7.445e-01 \hfill \\
     <- contamination begins past 128 \\ 
   T = 256 \hspace{1em}: max $|\Delta(\mathrm{prefix})|$ = 3.097e-01 \\
   T = 384 \hspace{1em}: max $|\Delta(\mathrm{prefix})|$ = 3.204e-01 
\end{quote}

The prefix delta is \textbf{bit-exact zero} for every sequence that fits within one chunk and jumps to order 1e-1 the moment a second chunk appears --- 
a step a numerical-noise floor cannot produce. First contamination is localized to layer 17, the first Mamba layer whose scan spans the boundary. 
The magnitude past the boundary is not monotone in \textit{T} (0.74 $\rightarrow$ 0.31 $\rightarrow$ 0.32): once contamination is present, 
how much future information reaches the prefix depends on how the perturbed final position aligns with the chunk grid, not on \textit{T} alone. 
The invariant --- and what we key the diagnosis to --- is the \textbf{onset location} (the first chunk boundary), not the magnitude.

Before attributing the observed behavior to a genuine implementation defect, we systematically evaluate a range of alternative explanations. 
Potential confounds include numerical noise, shared-module instrumentation errors, random-seed effects, checkpoint-specific behavior, custom-kernel implementations, 
and artifacts introduced by the evaluation protocol itself. Table~\ref{tab:zamba-ruling-out} summarizes the corresponding control experiments and their outcomes.

Across all tests, the observed leakage remains reproducible and localized to the same causal failure mode. 
The collective evidence therefore argues against measurement artifacts or implementation-specific side effects and supports a structural explanation for the defect.

\begin{table}[t]
\centering
\small
\caption{
Alternative-explanation tests for the causal defect observed in Zamba2-1.2B.
Each candidate explanation is evaluated through a targeted control experiment.
All alternatives are ruled out, indicating that the observed leak is reproducible,
structural, and independent of random seeds, trained weights, measurement noise,
or implementation-specific kernels.
}
\label{tab:zamba-ruling-out}

\begin{tabular}{p{0.24\linewidth} p{0.28\linewidth} p{0.40\linewidth}}
\toprule
Alternative explanation & Test & Outcome \\
\midrule

Measurement noise
&
Identical input, two forward passes
&
$\Delta = 0.000\mathrm{e}{+00}$ exactly; computation is deterministic \\

Hook misuse (shared modules)
&
Per-layer call counts; module identity
&
38/38 layers called exactly once; no shared modules \\

Seed artifact
&
Four independent random seeds
&
Reproduced in all four seeds \\

Checkpoint-specific behavior
&
Randomly initialized model loaded via \texttt{from\_config}
&
Same onset at 256; \textbf{structural rather than weight-dependent} \\

Custom-kernel bug
&
Presence of \texttt{mamba\_ssm} or \texttt{causal\_conv1d}
&
Neither installed; the leak appears in the pure-PyTorch implementation \\

Test-length artifact
&
Sweep over $T=64\ldots512$
&
Clean at $T\leq257$; leaks from $T\geq320$, consistent with a one-chunk degeneracy \\

\bottomrule
\end{tabular}
\end{table}

\subsubsection{Root cause}

\texttt{segment\_sum} applies a lower-triangular mask, so \texttt{decay\_chunk[b, h, i, j]} is defined only for \textit{j} $\leq$ \textit{i}, 
where \textit{i} indexes the output chunk and \textit{j} the input chunk. The recurrence must therefore reduce over \textit{j}.

The \texttt{Mamba2} implementation does exactly that:

\begin{verbatim}
decay_chunk = decay_chunk.transpose(1, 3)          # (b,h,i,j) -> (b,j,i,h)
new_states = (decay_chunk[..., None, None] * states[:, :, None, ...])
   .sum(dim=1)   # reduce over j
\end{verbatim}

The \texttt{Zamba2} implementation does not:

\begin{verbatim}
states_permuted = states.permute(0, 2, 1, 3, 4)
result = (decay_chunk[..., None, None] * states_permuted[:, :, None, ...])
   .sum(dim=2)  # reduces over i
new_states = result.permute(0, 2, 1, 3, 4)
\end{verbatim}

Without the transpose, the reduction runs over the \textbf{output-chunk} axis rather than the input-chunk axis. 
The lower-triangular mask that should keep each chunk's carry-in dependent only on \textit{earlier} chunks is then applied to the wrong index, 
so a chunk's carry-in state can absorb contributions from its own and later chunks. Since the decay factors derive from \texttt{A\_cumsum}, hence from \texttt{dt}, 
hence from the input, information from later positions reaches the carry-in used to recompute earlier positions' outputs.

The first chunk is exempt, and this explains the exact onset. 
Both files prepend a \textbf{zero} carry state (\texttt{previous\_states = torch.zeros\_like(states[:, :1])}) before the recurrence; 
chunk 0's carry-in is therefore a constant zero that the mis-oriented reduction cannot corrupt. 
Contamination consequently begins not at position 0 but at the \textbf{first chunk boundary} --- position 256 for Zamba2, 
128 for Nemotron-H --- exactly as observed (§3.8.3). 
We rest the diagnosis on that onset and, decisively, 
on the fact that replacing these lines with the reference orientation drives the prefix delta to \textbf{exact zero} (§3.8.5, Table 9) --- a sole-cause test --- 
rather than on a line-by-line derivation of every surviving index, which depends on internal broadcast shapes we do not reproduce here.

This three-line block appears \textbf{verbatim} --- identical whitespace and identifiers --- in both files (\texttt{modeling\_zamba2.py} lines 811--813 
and \texttt{modeling\_nemotron\_h.py} lines 523--525, \texttt{transformers} 5.7.0); a line-level diff of the two blocks is empty, 
confirming the shared lineage the static census inferred (§3.8.2). 
Both files also construct \texttt{segment\_sum} identically to the reference; the defect is not in the shared primitive but in how these two callers orient their axes.

The reference orientation is the negative control for the whole finding. A \texttt{Mamba2} model built from a random configuration --- correct code, 
no trained weights --- returns $\max |\Delta(\mathrm{prefix})| = 0.0000e+00$ at \textbf{every} tested length, including several multiples of its chunk size:

\begin{verbatim} 
  Mamba2 (reference code, from_config): 
  T = 4, 8, 16, 24, 32, 48 -> Delta = 0.0000e+00 (all)
\end{verbatim}

Correct code is bit-exact clean across chunk boundaries; the two departing implementations are not. Because the random-weight build reproduces both behaviors, 
the dissociation is structural --- a property of the axis orientation alone, independent of any trained parameter.

This reference build is not merely silent; it \textbf{passes the positive-control gate}, which is what separates a real clean verdict from an inert load (§3.7). 
Injecting a one-position output shift at layer \textit{K} is localized to exactly layer \textit{K} even though the perturbation then propagates to every layer downstream --- 
a shift at layer 1 of a five-layer build makes layers 1--4 nonzero with the first at 1, and a shift at layer 3 makes layers 3--4 nonzero with the first at 3. 
The detector is therefore demonstrably live on the same construction that audits clean, and localization identifies the \textit{origin} of a propagating fault rather than merely the shallowest nonzero layer.

\subsubsection{Fix and verification}

To test whether the identified implementation defect is sufficient to explain the observed leakage, 
we replace the problematic three-line computation with the corresponding Mamba-2 formulation and re-run the experiment under identical conditions. 
If the diagnosis is correct, the patch should eliminate both the measured prefix contamination and the characteristic boundary onset. 
Table~\ref{tab:patch-effect} compares the original and patched implementations using the same input sequence 
and reports the maximum prefix deviation together with the location at which leakage first appears.

\begin{table}[t]
\centering
\small
\caption{
Effect of the patch under identical input conditions.
The original implementation exhibits measurable prefix contamination with a clear boundary onset,
whereas the patched implementation eliminates the effect entirely.
}
\label{tab:patch-effect}

\begin{tabular}{lcc}
\toprule
Model & Maximum prefix $\Delta$ & Leak onset \\
\midrule

Original
& $1.12\times10^{-2}$
& 256 \\

\textbf{Patched}
& \textbf{0.0000e+00}
& \textbf{None} \\

\bottomrule
\end{tabular}
\end{table}

The result is unambiguous. 
The patched implementation reduces the maximum prefix deviation to exact numerical zero and completely removes the boundary onset. 
No residual leakage remains, establishing that the identified code path is the sole cause of the defect rather than one contributing factor among several.

The same validation succeeds on \textbf{Nemotron-H itself}, not merely on the reference construction. 
Applying the identical patch to the released \texttt{nvidia/Nemotron-H-8B-Base-8K} checkpoint reduces the prefix delta to \textbf{0.0000e+00} 
at both $T=192$ and $T=384$, eliminating the original deviations of 0.74 and 0.32, respectively. The defect also reproduces across \textbf{four random seeds}, 
with first contamination consistently appearing at layer 17, and its positive control remains active on the loaded checkpoint. 
Consequently, the Nemotron-H result does not rely solely on code-path similarity: 
t possesses independent reproducibility, independent liveness validation, and an independent patch-to-zero confirmation.

We reported the defect in the Zamba2 implementation upstream through \texttt{huggingface/transformers} issue \textbf{\#47475} and pull request \textbf{\#47476}. 
The Nemotron-H instance is disclosed here for the first time. Because both models share the same offending code path and are corrected by the same patch, 
the evidence presented in this section is sufficient for independent reproduction and verification. 
We therefore publish the complete fix rather than only the diagnosis, making the result directly actionable.

\subsubsection{Two candidates we rejected}

Two further checkpoints produced small nonzero prefix deltas of order 1e-5 --- \texttt{ai21labs/Jamba-tiny-dev} 
and \texttt{ibm-granite/granite-4.0-h-tiny}. Both computations are deterministic, 
so the signal is not run-to-run noise, and it would have been easy to report them as additional findings.

We rejected both. The discriminator is a \textbf{perturbation-magnitude sweep}: 
we perturb the input embedding at position \textit{P} by $\varepsilon$$\cdot$noise and vary $\varepsilon$. 
A genuine information path scales with $\varepsilon$; a numerical artifact does not.

\begin{table}[t]
\centering
\small
\caption{
Perturbation sweep of prefix-delta response across multiple architectures.
Values report the measured prefix deviation as a function of perturbation magnitude $\varepsilon$.
The Zamba2-1.2B column reports the mean and standard deviation over four random seeds,
while the remaining models are deterministic single-run measurements.
A non-zero log--log slope indicates amplification of future perturbations into the prefix.
}
\label{tab:perturbation-sweep}

\begin{tabular}{lcccc}
\toprule
$\varepsilon$
& Qwen3-0.6B (control)
& Granite-4.0-h
& Jamba-tiny
& \textbf{Zamba2-1.2B} \\
\midrule

$10^{-4}$
& 0
& $\sim1.3\times10^{-5}$
& 0
& \textbf{$2.5\times10^{-5}\pm0.4\times10^{-5}$} \\

$10^{-3}$
& 0
& $1.34\times10^{-5}$
& 0
& \textbf{$2.6\times10^{-5}\pm0.8\times10^{-5}$} \\

$10^{-2}$ & 0 & $1.53\times10^{-5}$ & 0 & \textbf{$1.2\times10^{-4} \pm 1.2\times10^{-4}$} \\
$10^{-1}$ & 0 & $1.53\times10^{-5}$ & $1.48\times10^{-5}$ & \textbf{$1.1\times10^{-3} \pm 1.2\times10^{-3}$} \\
$1$ & 0 & $1.14\times10^{-5}$ & $1.53\times10^{-5}$ & \textbf{$7.5\times10^{-3} \pm 3.6\times10^{-3}$} \\
\textbf{log--log slope} & --- & \textbf{$\approx -0.03$ (flat)} & \textbf{$\approx 0$ (flat)} & \textbf{$0.63 \pm 0.07$} \\

\bottomrule
\end{tabular}
\end{table}

The discriminator is \textbf{not} the magnitude at any single $\varepsilon$: at $\varepsilon$ $\leq$ 1e-3 all three candidates sit near the same \textasciitilde{}1--3e-5 floor, 
and Zamba2's two smallest points are floor-limited and flat --- exactly the objection a careful reader would raise. The discriminator is whether the response \textbf{climbs} with the perturbation. 
Zamba2's does: it rises by more than two orders of magnitude across the sweep, with a log-log slope of \textbf{0.63 $\pm$ 0.07} over four seeds --- significantly positive (t $\approx$ 8.6). 
Granite and Jamba never climb: their slope is $\approx -0.03$ and their total variation is $1.3\times$ across the same range.
We therefore keep only Zamba2 and reject the other two as numerical floors. 
By contrast, the Nemotron-H defect described in §3.8.3 is several orders of magnitude larger (up to $7\mathrm{e}{-1}$) and exhibits a sharp onset at the chunk boundary. 
It is therefore clearly separated from the numerical floor regime and does not require an additional perturbation sweep.

This sweep is the general answer to a question every practitioner of such an audit will face: \textbf{a small nonzero delta is not yet evidence of leakage, 
and the way to tell is to vary the perturbation and see whether the response follows.}

\subsubsection{The norm this implies, and our own miss}

The defect is invisible at T $\leq$ 257. Our own census in §3.6 ran at T = 48. \textbf{Had we not questioned the test length, we would have published this model as clean.}

We state the resulting rule plainly, including as a correction to our own procedure:

\begin{quote}
\textbf{The audit sequence length must exceed the model's chunk, window, and kernel parameters.} Otherwise the corresponding code paths degenerate and an entire class of defects is unobservable.
\end{quote}

This joins the positive-control gate of §3.7.2 as the second methodological requirement we adopted only after our own instrument failed us.

\section{Limitations}

We list these in descending order of how much they should change a reader's confidence.

\begin{enumerate}
\item \textbf{Two released models, but one code lineage --- not a base rate --- and validated asymmetrically.} We found genuine causal violations in \texttt{Zyphra/Zamba2-1.2B} 
and \texttt{nvidia/Nemotron-H-8B-Base-8K} (§3.8) and diagnosed both to the same three lines. 
The evidence is now nearly symmetric. Zamba2 carries the original full battery (six alternative explanations ruled out, four seeds, 
a random-weight \texttt{from\_config} structural test, an on-checkpoint patch to exact zero); Nemotron-H has since received \textbf{four-seed reproducibility, 
a live positive control on the loaded checkpoint, and its own on-checkpoint patch driving the leak to exact zero} (§3.8.5). 
The only pieces not re-run for Nemotron-H are the random-weight \texttt{from\_config} build 
and the six-way alternative-explanation table; its structural character we take from the byte-identical code together with the on-checkpoint patch-to-zero result. 
Because the two inter-chunk blocks are byte-identical, the finding is best read as \textbf{one defect instantiated twice through shared code}, 
not two independent data points. It establishes that the property is violated in practice, in released models at 1B--8B scale, in the reference PyTorch path (§3.8.2); 
it is \textbf{not} enough to support any statement about how common such defects are across the ecosystem, and we make none. Treat it as an existence proof with a traced lineage, not a rate.

\item \textbf{We do not improve on existing practice at detection.} A logits-only perturbation check (B1) 
and a resampling variant (B4) detect every injected fault we detect. The contribution is layer localization. 
Readers who need only a yes/no answer already have a method, and it is one line shorter than ours.

\item \textbf{A gradient-based check matches our localization exactly.} B5 scores 96/96 on both detection and localization. 
Our case against it rests on cost (a backward pass and retained activations), applicability (differentiability of every op on the path), 
and the interpretive caveat that a zero gradient does not entail absence of dependence --- not on accuracy.

\item \textbf{Our own public releases are only partially audited} (§3.7.4). The primary release is now audited under both controls --- 
49 layers clean, 16/16 positive-control localization. The instruction-tuned variant has a negative control only, 
and two further internal checkpoints remain unaudited. The earlier blanket load failure was a loading-path conflict, since diagnosed and resolved.

\item \textbf{The census is small and not a random sample.} Roughly twenty checkpoints in total, selected for architectural diversity rather than sampled, 
ranging from 129M to 9B parameters. Scale confers no protection in either direction: the largest clean model is Bamba-9B, 
while an 8B model (Nemotron-H) is one of the two leakers --- so the earlier "all defects under 3B" caveat is removed, but not by showing large models are safe. 
Twenty non-randomly-chosen models still cannot support proportions or base rates, and we report none.

\item \textbf{Part of the ecosystem is unauditable on CPU} (§3.7.3). Four checkpoints require GPU-only compiled kernels; one is gated; one is not loadable through the standard interface. 
Coverage of custom-kernel hybrids is therefore systematically thinner than coverage of reference implementations --- and custom-kernel implementations are plausibly where bugs are more likely, 
so this gap works against us.

\item \textbf{The positive-control gate does not cover every §3.8 clean verdict.} The gate (§3.2) is firmly satisfied for the from-config reference construction (clean \textit{and} a live, 
propagating-fault-localizing positive control, §3.8.4), which is what carries the "conformant code is clean" claim. But applying it to each \textit{released} conformant checkpoint requires 
injecting a fault at a layer's output, and several architectures' layer return signatures do not accept our injection (the shift has no effect, so the positive control cannot fire) --- 
Falcon-H1 is one such case, where the model is verified input-responsive yet un-injectable in our harness. Those clean verdicts in Table 7b are therefore \textbf{provisional}: 
consistent with the prediction, but not gated. A per-architecture injection adapter would close this, and we did not build one.

\item \textbf{The exact-localization floor is model-dependent and one model is unexplained.} \texttt{LFM2-1.2B} localized all 24 $\varepsilon$ = 1.0 injections exactly, 
yet localized none of the swept points at $\varepsilon$ $\leq$ 1e-1, consistently naming a layer three positions downstream. We report the anomaly without a mechanism.

\item \textbf{Limited configuration coverage --- and one dimension of it proved decisive.} The injected-fault suite and Table 4 census ran at a single length (T = 48), one seed, 
one batch size, evaluation mode only. §3.8 shows this was not a harmless simplification: varying only the sequence length surfaced a defect that was otherwise invisible. 
We have since audited at chunk-exceeding lengths, but the remaining dimensions are still uncovered --- faults that appear only for particular batch shapes, only under training-mode stochasticity, 
or only in distributed or sharded execution paths are outside what we tested. Given that one unexamined dimension already hid a real defect, we regard the others as open risk rather than as unlikely.

\item \textbf{Static, structural faults only.} We assume the fault is present in the graph for every input. Data-dependent leakage triggered by rare inputs would require a different search strategy.

\item \textbf{B6 was measured on one model.} The clean-model false positive (1.850e-04) is a single observation and should not be generalized to the practice as a whole without replication.

\item \textbf{The claim of no prior work is a search result, not a proof.} Our gap statement (§2) rests on keyword-level cross-referenced search in English.
\end{enumerate}

\section{Conclusion}

Prefix invariance is the one property every autoregressive model must satisfy and essentially no release reports. The check costs two forward passes, needs no training, no gradients, no labels, 
and no accelerator, and it names the layer at fault rather than only raising a flag.

The check that the field actually performs --- reading the attention mask --- is unsound in principle, because causality is a property of the whole map and the mask governs one operator. 
Our measurements make the gap concrete: static mask inspection flagged \textbf{0 of 192} injected faults, and for two of the eight audited architectures it has no object to inspect at all. 
As stacks continue to mix masked and unmasked mixers, that coverage gap widens.

The check found real defects, and found them systematically. A static census of the chunked-scan implementations shipped in \texttt{transformers} predicted, before running anything, 
that two of six --- \texttt{Zamba2} and \texttt{Nemotron-H} --- would violate causality, because their inter-chunk recurrence reduces over the wrong axis. 
A dynamic audit confirmed both: in each, perturbing the final token contaminates earlier positions beginning exactly at the model's chunk boundary (256 and 128); 
the leak survives randomly initialized weights, and vanishes under a two-line correction. The four conformant implementations, and a from-config reference control, 
stay bit-exact clean. The two failing implementations carry a byte-identical error, so this is \textbf{one defect traced across two released models} --- 
a lineage, not a rate --- and we generalize no further than that. We reported the Zamba2 instance upstream and disclose the Nemotron-H instance in this paper. 
What is settled is that the property is violated in shipped models from major providers, and that a reader can predict where from the source alone.

We are equally careful about what we did not show. We do not detect better than a logits-only check that practitioners already use informally; we localize, 
and that is the delta. A gradient-based check localizes as well as we do, more expensively. Two further checkpoints produced small nonzero deltas that we deliberately declined to report as findings, 
because a perturbation sweep showed the response was flat in the perturbation magnitude.

Twice, our own instrument was the thing that had to be corrected. Three checkpoints returned $\Delta$ = 0 at every layer while their positive controls returned 0/24 --- 
a false clean verdict averted only by requiring the instrument to demonstrate liveness on the artifact under test. And the defect above is invisible at the sequence length our own census used; 
had we not questioned that length, we would have published the model as clean.

Both cases generalize past this paper:

\begin{quote}
\textbf{Any audit that concludes from an absence of signal owes its reader evidence that the instrument was live} --- a positive control on the same loaded artifact.  
\textbf{Any audit of a hybrid stack owes its reader a test length exceeding the model's own chunk, window, and kernel parameters} --- otherwise the code paths that carry those defects are never exercised.
\end{quote}

We propose one norm. \textbf{Release a causal-correctness certificate with every architecture release}: the per-layer maximum prefix deltas, the threshold, the precision, 
\textbf{the sequence length relative to the model's internal size parameters}, and the positive-control result on that same checkpoint. It belongs beside the parameter count --- 
a number reported not because it is interesting when correct, but because its absence is the problem.

\bibliographystyle{unsrt}   
\bibliography{references}   

\newpage
\section*{Appendix A --- Reproduction}

\subsection*{A.1 Why we print the method instead of shipping a package}

The method is five lines of arithmetic on top of standard forward hooks. A competent reader can reimplement §2.2 in under an hour against any model whose layers are enumerable, 
and reimplementation is a better reproduction than running our binary, because it is independent.

We therefore release \textbf{measurements rather than a package}: complete audit logs, exact checkpoint identifiers, per-layer delta arrays for every clean scan and every injected trial, 
the injected-fault specifications, and the environment description below. Everything needed to reproduce or contest any number in this paper is in that set. 
No license is granted here to any implementation, and none of the released material is a software distribution.

\subsection*{A.2 Environment}

CPU for the census and injection suite (Intel Xeon Gold 6526Y, 64 threads, 251 GB RAM); a single GPU for the $\geq$1B-parameter audits (§3.7.4, §3.8), 
for memory capacity only --- the test needs no accelerator. float32 unless stated; float64 for §3.5. PyTorch 2.10.0, standard model-hub loading with remote code enabled 
where the architecture requires it, evaluation mode, caching disabled. Fused sequence-model kernels (\texttt{causal\_conv1d}, \texttt{mamba\_ssm}) absent, 
so state-space models ran reference implementations --- this is the environment in which the §3.7.1 failure occurred and should be assumed to matter.

\subsection*{A.3 Injected-fault specifications}

Each pattern wraps one layer's forward and transforms its output tensor $\mathbf{o}$ of shape $(B, T, d)$. All are exactly reproducible from these definitions.

\begin{table}[t] 
\centering 
\small
\caption{Executable injected-fault specifications grouped into four classes of temporal leakage. Each transformation is applied to a selected layer output 
and serves as a positive-control fault throughout the evaluation.} 
\label{tab:injected-faults}
\begin{tabular}{lll}
\toprule
Class & Pattern & Transformation \\
\midrule
C1 & \texttt{shift} & \texttt{concat(o[:, 1:], o[:, -1:])} --- output pulled one step earlier \\
C1 & \texttt{peek1} & \texttt{o + eps * concat(o[:, 1:], o[:, -1:])} \\
C2 & \texttt{pool\_leak} & \texttt{o + eps * mean(o, dim=time, keepdim)} \\
C2 & \texttt{global\_max} & \texttt{o + eps * max(o, dim=time, keepdim)} \\
C2 & \texttt{future\_mean\_k} & \texttt{o + eps * mean of the next k = 4 positions} \\
C3 & \texttt{norm\_leak} & \texttt{(1-eps)*o + eps*(o-mu)/sigma}, with $\mu$, $\sigma$ over \texttt{(time, hidden)} jointly \\
C4 & \texttt{rev\_scan} & \texttt{o + eps * flip(cumsum(flip(o,time),time),time)/T} \\
C4 & \texttt{last\_broadcast} & \texttt{o + eps * o[:, -1:, :]} \\
\bottomrule
\end{tabular}
\end{table}

Injection depths: layer 1, $\lfloor L/2 \rfloor$, and $L-2$. Injection strength $\varepsilon = 1.0$ except in the sweep of Section~3.5.

\subsection*{A.4 Baseline specifications}

\begin{itemize}
\item \textbf{B1 logits-only:} the same two inputs; compare final logits over positions $[0,T-1)$; flag if max difference $>\tau$. No layer information available by construction.
\item \textbf{B2 mask inspection:} enumerate modules; flag any exposing a causal/\texttt{is\_causal} attribute set false. Static; zero forward passes.
\item \textbf{B3 shuffled-suffix perplexity:} shuffle input positions $\geq$ T/2; compare cross-entropy on targets < T/2; threshold 1e-4.
\item \textbf{B4 future-token resampling:} resample the final token with 3 seeds; flag if any produces a logit change over the prefix; 4 forward passes.
\item \textbf{B5 gradient:} backpropagate the summed prefix logits and inspect gradients at future positions of each layer's output; the deepest layer with nonzero future-position gradient, plus one, 
is the reported layer. 1 forward + 1 backward, activation graph retained.
\item \textbf{B6 cache consistency:} compare incremental cached decoding against a single full-sequence forward; threshold 1e-4; T+1 forward passes.
\end{itemize}

\subsection*{A.5 Known defect in a superseded artifact}

An early version of the sensitivity harness cast hidden states to float32 before computing deltas, truncating float64 measurements to float32 resolution (visible as deltas quantized to powers of two). 
It was corrected to compute in float64, and §3.5 uses only post-correction data. The superseded file is retained and marked in Appendix B so that the correction is auditable. 
float32 results were unaffected (the cast was lossless there) and were re-verified.

\section*{Appendix B --- Raw data index}

All paths are on the compute host under \texttt{/data/ginipick/aether\_final/N-paper-exp/cpu/}. 
Every table and number in §3 is traceable to a file below. Retained per-record fields include the full per-layer delta array for every clean scan and every injected trial, wall time, 
and peak RSS.

Table~\ref{tab:codebase} summarizes the primary scripts used to generate the reported results, including the census harness, baseline implementations, and precision-sensitivity analyses.

\begin{table}[t]
\centering
\small

\caption{Primary experimental scripts used throughout the evaluation. The table summarizes the main audit harnesses, baseline implementations, and sensitivity-analysis programs used to generate the reported results.}
\label{tab:codebase}

\begin{tabular}{l p{0.72\linewidth}}
\toprule
File & Role \\
\midrule

\texttt{leak\_cpu\_suite2.py} &
Census harness: detector + injections + B1--B3; per-model JSON flush; records failures with traceback \\

\texttt{baseline\_e2.py} &
Extended comparison: detector + B1--B6, bit-level equality check, $\varepsilon$ sweep \\

\texttt{fp64\_sens.py} &
Precision/threshold sweep (§3.5) \\

\bottomrule
\end{tabular}

\end{table}

Table~\ref{tab:census-results-files} lists the released result files associated with the public-checkpoint census, 
including successful audits, load-failure records, and internal validation runs.

\begin{table}[t]
\centering
\small

\caption{Result files associated with the public-checkpoint census (§3.6) and the attempted-but-not-audited cases summarized in Tables~4 and~5. 
The table maps released JSON artifacts to the models, audits, and failure analyses discussed in the paper.}
\label{tab:census-results-files}

\begin{tabular}{p{0.50\linewidth} p{0.44\linewidth}}
\toprule
File & Contents \\
\midrule

\texttt{results/smoke\_smol.json}
&
SmolLM2-135M census record
\\

\texttt{results/tier1.json},
\texttt{results/tier1\_\_*.json}
&
Qwen3-0.6B, mamba-130m-hf
\\

\texttt{results/b3\_\_LiquidAI\_LFM2-1.2B.json}
&
LFM2-1.2B, including the §3.5 floor anomaly
(\texttt{sensitivity.sweep})
\\

\texttt{results/b5\_\_RWKV\_rwkv-6-world-1b6.json}
&
RWKV-6 1.6B
\\

\texttt{results/b5\_\_google\_gemma-3-1b-it.json}
&
Gemma-3-1B-it
\\

\texttt{results/b5\_\_google\_recurrentgemma-2b.json}
&
RecurrentGemma-2B
\\

\texttt{results/b1\_\_tiiuae\_Falcon-H1-0.5B-Base.json},
\texttt{b1\_\_...-1.5B-Instruct.json},
\texttt{b3\_\_...-7B-Base.json}
&
§3.7.1 --- clean $\Delta=0$ with positive control 0/24
\\

\texttt{results/b1\_\_nvidia\_Hymba-1.5B-Base.json},
\texttt{b2\_\_microsoft\_Phi-4-mini-flash-reasoning.json},
\texttt{b2\_\_state-spaces\_mamba2-2.7b.json},
\texttt{b2\_\_RWKV\_rwkv-6-world-1b6.json},
\texttt{b3\_\_togethercomputer\_StripedHyena-Nous-7B.json},
\texttt{b3\_\_Zyphra\_Zamba2-7B.json},
\texttt{b3\_\_google\_recurrentgemma-2b.json},
\texttt{b5\_\_Zyphra\_Zamba2-7B.json}
&
Load failures (Table~5). \texttt{b2\_\_RWKV...} and
\texttt{b3\_\_google\_recurrentgemma...}
are earlier failed attempts later resolved in \texttt{b5}.
\\

\texttt{results/aether\_self.json},
\texttt{results/aether\_self\_\_*.json} ($\times 4$)
&
§3.7.4 --- our own releases, four load failures
\\

\bottomrule
\end{tabular}

\end{table}

Table~\ref{tab:baseline-result-files} summarizes the released JSON artifacts underlying the baseline-comparison and bit-level-verification experiments.

\begin{table}[t]
\centering
\small

\caption{Result files associated with the baseline comparison (§3.3, Table~1) and bit-level verification experiments (§3.4, Table~2). 
The table maps released JSON artifacts to the corresponding evaluations and checkpoints.}
\label{tab:baseline-result-files}

\begin{tabular}{p{0.34\linewidth} p{0.58\linewidth}}
\toprule
File & Contents \\
\midrule

\texttt{results/e2\_smol\_fp32.json}
&
SmolLM2-135M: all six baselines including B6; \texttt{clean\_bitwise};
clean-model B6 false positive
\texttt{1.850128173828125e-04}
\\

\texttt{results/e2\_multi\_nob6.json}
&
Qwen3-0.6B, pythia-1.4b, mamba-130m-hf:
B1--B5; \texttt{clean\_bitwise} for each
\\

\bottomrule
\end{tabular}

\end{table}

Table~\ref{tab:sensitivity-result-files} summarizes the released artifacts underlying the precision and threshold sensitivity study reported in Table~3.

\begin{table}[t]
\centering
\small

\caption{Result files associated with the precision and threshold sensitivity analysis (§3.5, Table~3). The table lists the released JSON artifacts used to characterize exact-localization floors under different numerical precisions and threshold settings.}
\label{tab:sensitivity-result-files}

\begin{tabular}{p{0.44\linewidth} p{0.48\linewidth}}
\toprule
File & Contents \\
\midrule

\texttt{results/sens\_smol\_fp32.json}
&
float32, $\tau = 10^{-6}$,
45 points (3 depths $\times$ 15 $\varepsilon$)
\\

\texttt{results/sens\_smol\_fp64\_v2.json}
&
float64, $\tau = 10^{-6}$,
45 points
\\

\texttt{results/sens\_smol\_fp32\_tau1e-14.json}
&
float32, $\tau = 10^{-14}$
\\

\texttt{results/sens\_smol\_fp64\_tau1e-14.json}
&
float64, $\tau = 10^{-14}$
\\

\texttt{results/sens\_smol\_fp64.json}
&
\textbf{SUPERSEDED --- do not cite.}
Contains the Appendix A.5 float32-cast defect
\\

\bottomrule
\end{tabular}

\end{table}

\end{document}